\documentclass[10pt,twocolumn,letterpaper]{article}

\usepackage{iccv}
\usepackage{times}
\usepackage{epsfig}
\usepackage{graphicx}
\usepackage{amsmath}
\usepackage{amssymb}

\usepackage{bm}
\usepackage{bbm}
\usepackage{multirow}
\usepackage{graphicx}
\usepackage{subfigure}
\newcommand{\model}{\textsc{NMTree} }
\newcommand{\modelend}{\textsc{NMTree}}

\usepackage[pagebackref=true,breaklinks=true,letterpaper=true,colorlinks,bookmarks=false]{hyperref}

\iccvfinalcopy

\begin{document}

%%%%%%%%% TITLE
\title{Learning to Assemble Neural Module Tree Networks for Visual Grounding}

\author{Daqing Liu$^{1}$ ~~~~~~~Hanwang Zhang$^{2}$ ~~~~~~~Feng Wu$^1$ ~~~~~~~Zheng-Jun Zha$^1$\thanks{\scriptsize{Corresponding Author.}}\\
$^1$University of Science and Technology of China,
	$^2$Nanyang Technological University \\\tt\small{
	liudq@mail.ustc.edu.cn, hanwangzhang@ntu.edu.sg, fengwu@ustc.edu.cn, zhazj@ustc.edu.cn}
}

\maketitle

\begin{abstract}
Visual grounding, a task to ground (\textit{i.e.}, localize) natural language in images, essentially requires composite visual reasoning. However, existing methods over-simplify the composite nature of language into a monolithic sentence embedding or a coarse composition of subject-predicate-object triplet.
In this paper, we propose to ground natural language in an intuitive, explainable, and composite fashion as it should be. In particular, we develop a novel modular network called Neural Module Tree network (\modelend) that regularizes the visual grounding along the dependency parsing tree of the sentence, where each node is a neural module that calculates visual attention according to its linguistic feature, and the grounding score is accumulated in a bottom-up direction where as needed. \model disentangles the visual grounding from the composite reasoning, allowing the former to only focus on primitive and easy-to-generalize patterns.
To reduce the impact of parsing errors, we train the modules and their assembly end-to-end by using the Gumbel-Softmax approximation and its straight-through gradient estimator, accounting for the discrete nature of module assembly.
Overall, the proposed \model consistently outperforms the state-of-the-arts on several benchmarks. Qualitative results show explainable grounding score calculation in great detail.
% Visual grounding, a task to ground (i.e., localize) natural language in images, essentially requires composite visual reasoning. However, existing methods over-simplify the composite nature of language into a monolithic sentence embedding or a coarse composition of subject-predicate-object triplet. In this paper, we propose to ground natural language in an intuitive, explainable, and composite fashion as it should be. In particular, we develop a novel modular network called Neural Module Tree network (NMTree) that regularizes the visual grounding along the dependency parsing tree of the sentence, where each node is a neural module that calculates visual attention according to its linguistic feature, and the grounding score is accumulated in a bottom-up direction where as needed. NMTree disentangles the visual grounding from the composite reasoning, allowing the former to only focus on primitive and easy-to-generalize patterns. To reduce the impact of parsing errors, we train the modules and their assembly end-to-end by using the Gumbel-Softmax approximation and its straight-through gradient estimator, accounting for the discrete nature of module assembly. Overall, the proposed NMTree consistently outperforms the state-of-the-arts on several benchmarks. Qualitative results show explainable grounding score calculation in great detail.
\end{abstract}
\section{Introduction}
Visual grounding (\textit{a.k.a.}, referring expression comprehension) aims to localize a natural language description in an image.
It is one of the core AI tasks for testing the machine comprehension of visual scene and language~\cite{krahmer2012computational}.
Perhaps the most fundamental and related grounding system for words is object detection~\cite{ren2015faster} (or segmentation~\cite{he2017mask}): the image regions (or pixels) are classified to the corresponding word of the object class.
Despite their diverse model architectures~\cite{liu2018deep}, their sole objective is to calculate a grounding score for a visual region and a word, measuring the semantic association between the two modalities.
Thanks to the development of deep visual features~\cite{he2016deep} and language models~\cite{mikolov2010recurrent}, we can extend the grounding systems from fixed-size inventory of words to open-vocabulary~\cite{hu2018learning} or even descriptive and relational phrases~\cite{zhang2017visual, plummer2017phrase}.

\begin{figure}[t]
\centering
\subfigure[Holistic]{
\includegraphics[width=0.484\linewidth]{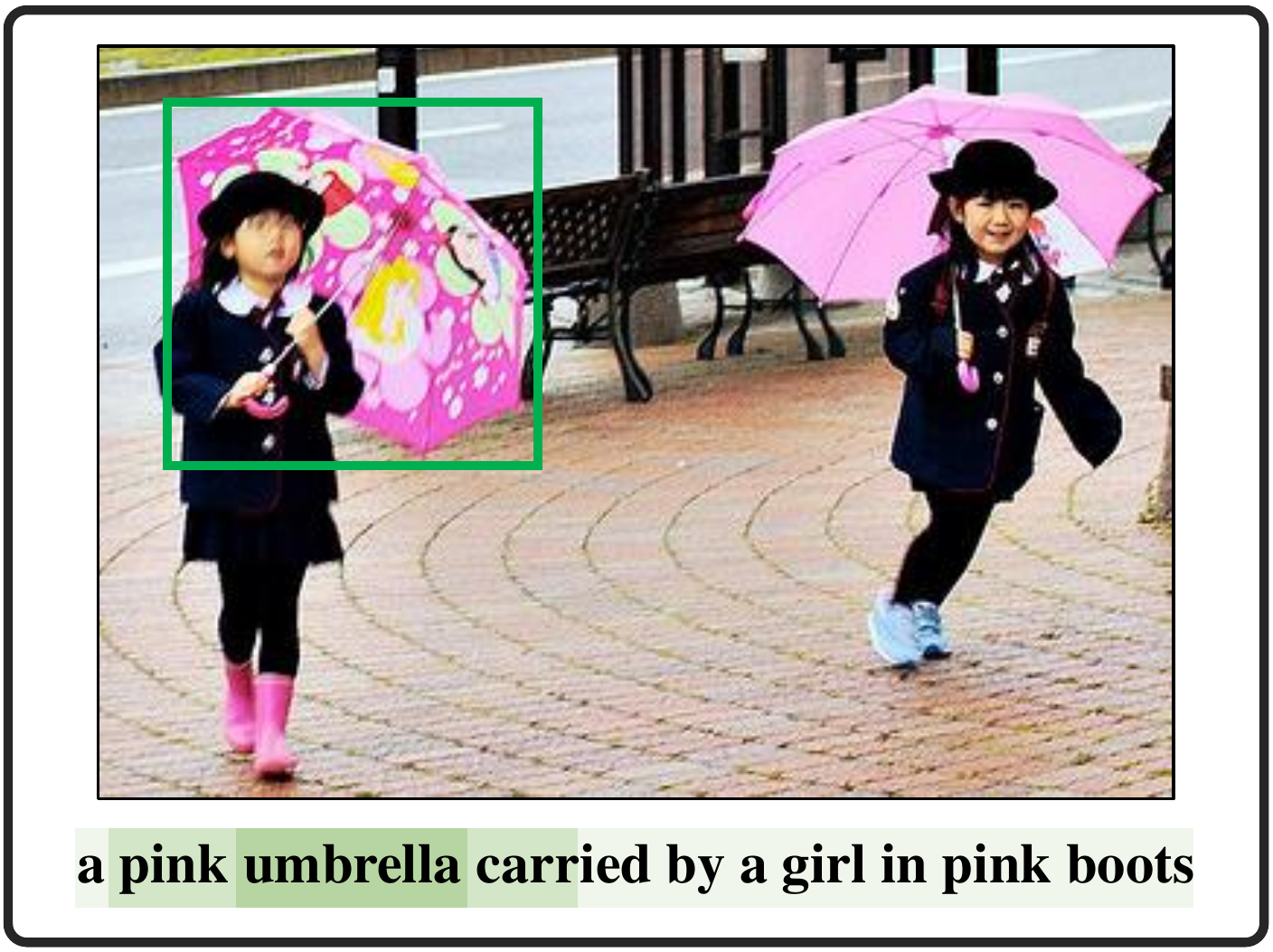}
\label{fig:1a}
}%
\subfigure[Triplet]{
\includegraphics[width=0.484\linewidth]{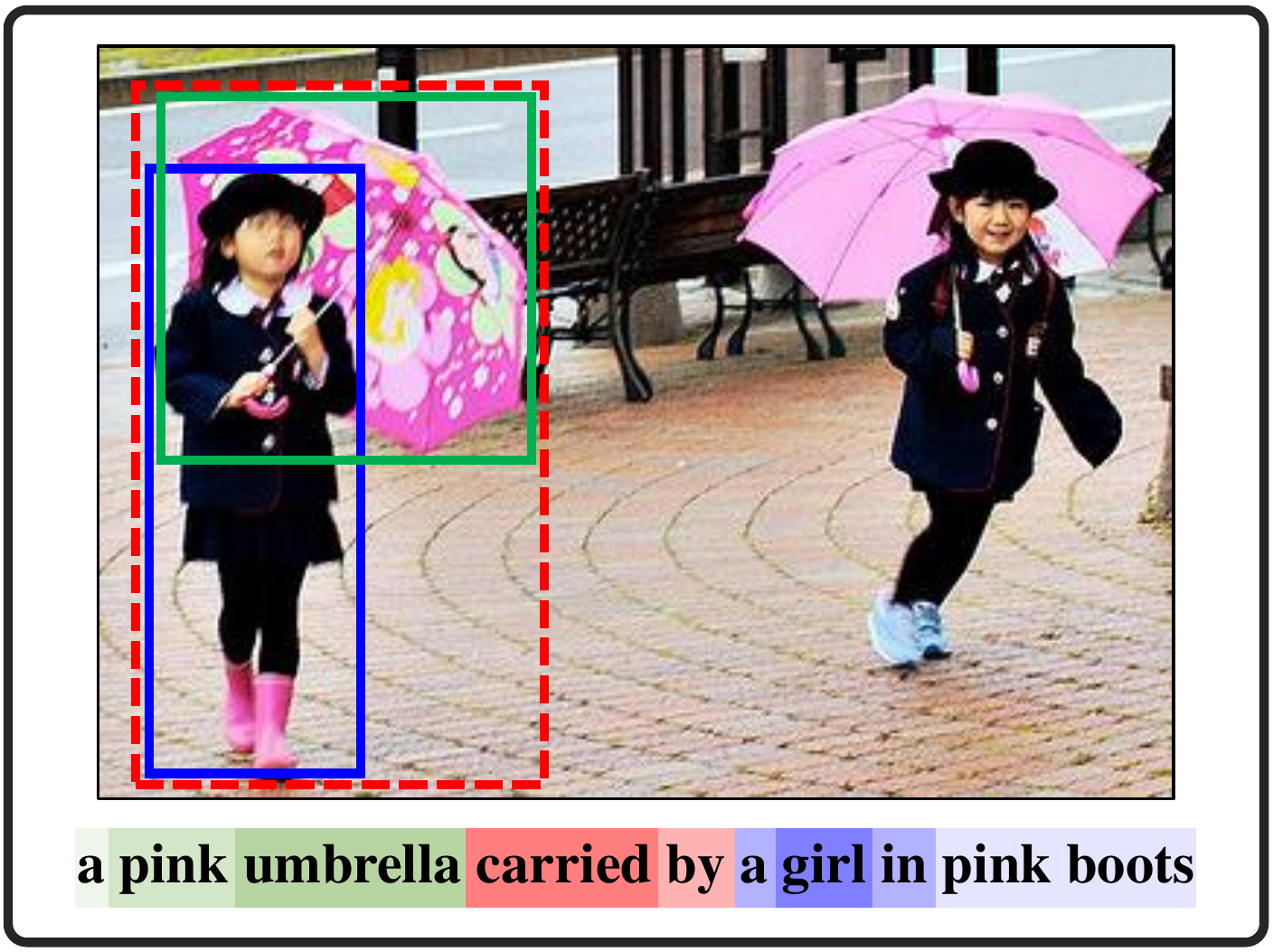}
\label{fig:1b}
}%
\vspace{-5pt}
\subfigure[\model]{
\includegraphics[width=0.99\linewidth]{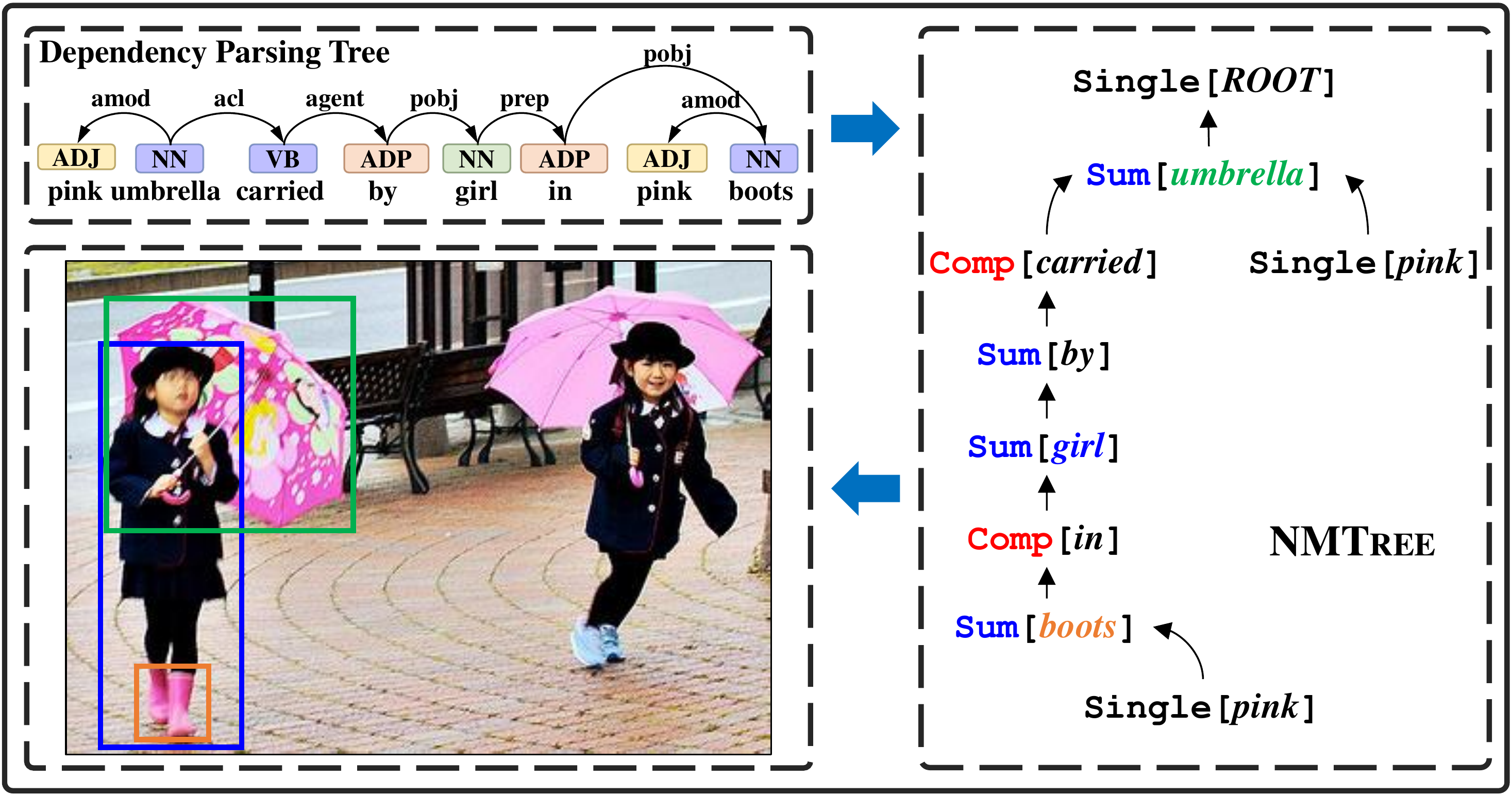}
\label{fig:1c}
}%
\centering
\vspace{1pt}
\caption{Existing grounding models are generally (a) holistic or (b) coarsely composite. Words in gradient colors indicate word-level attentions. (c) The proposed \model is based on dependency parsing tree and offers explainable grounding in great detail. Word color corresponds to image regions.}
\vspace{-8pt}
\label{fig:1}
\end{figure}

However, grounding complex language sentences, \eg, ``a pink umbrella carried by a girl in pink boots'', is far different from the above word or phrase cases. For example, given the image in Figure~\ref{fig:1}, for us humans, how to localize the ``umbrella''? One may have the following reasoning process: 1) Identify the referent ``umbrella'', but there are two of them. 2) Use the contextual evidence ``carried by a girl'', but there are two girls. 3) By using more specific evidence ``in pink boots'', localize the ``girl'' in the last step. 4)~Finally, by accumulating the above evidences, localize the target ``umbrella''.

Unfortunately, existing visual grounding methods generally rely on 1) a single monolithic score for the whole sentence~\cite{mao2016generation, yu2016modeling, luo2017comprehension, yu2017joint} (Figure~\ref{fig:1a}), or 2) a composite score for subject, predicate, and object phrases~\cite{hu2017modeling, yu2018mattnet} (Figure~\ref{fig:1b}). Though some of them adopt the word-level attention mechanism~\cite{luong2015effective} to focus on the informative language parts, their reasoning is still coarse compared to the above human-level reasoning. More seriously, such coarse grounding scores are easily biased to learn certain vision-language patterns but not visual reasoning, \eg, if most of the ``umbrellas'' are ``carried by people'' in the dataset, the score may not be responsive to other ones such as ``people under umbrella stall''. Not surprisingly, this problem has been repeatedly discovered in many end-to-end vision-language embedding frameworks used in other tasks such as VQA~\cite{johnson2017clevr} and image captioning~\cite{lu2018neural}.

In this paper, we propose to exploit the Dependency Parsing Trees (DPTs)~\cite{chen2014fast} that have already offered an off-the-shelf schema for the composite reasoning in visual grounding. Specifically, to empower the visual grounding ability by DPT, we propose a novel neural module network: Neural Module Tree (\modelend) that provides explainable grounding scores in \emph{great} detail. As illustrated in Figure~\ref{fig:1c}, we transform a DPT into \model by assembling three primitive module networks: \texttt{Single} for leaves and root, \texttt{Sum} and \texttt{Comp} for internal nodes (detailed in Section~\ref{sec:3.3}). Each module calculates a grounding score, which is accumulated in a bottom-up fashion, simulating the visual evidence gained so far. For example in Figure~\ref{fig:1c}, $\texttt{Comp}[carried]$ receives the scores gained by $\texttt{Sum}[by]$ and then calculates a new score for the region composition, meaning ``something is carried by the thing that is already grounded by the `by' node''. Thanks to the fixed reasoning schema, \model disentangles the visual perception from the composite reasoning to alleviate the unnecessary vision-language bias~\cite{yi2018neural}, as the primitive modules receive consistent training signals with relatively simpler visual patterns and shorter language constitutions.

One maybe concerned by the potential brittleness caused by DPT parsing errors that impact the robustness of the module assembly, as discovered in most neural module networks applied in practice~\cite{hu2017learning, cao2018visual}. We address this issue in three folds: 1) the assembly is simple. Except for \texttt{Single} that is fixed for leaves and root, only \texttt{Sum} and \texttt{Comp} are to be determined at run-time; 2) \texttt{Sum} is merely an Add operation that requires no visual grounding; 3) we adopt the recently proposed Gumbel-Softmax (GS) approximation~\cite{jang2016categorical} for the discrete assembly approximation. During training, the forward pass selects the two modules by GS sampler in a ``hard'' discrete fashion; the backward pass will update all possible decisions by using the straight-through gradient estimator in a ``soft'' robust way. By using the GS strategy, the entire \model can be trained end-to-end without any additional module layout annotations.

We validate the effectiveness of \model on three challenging visual grounding benchmarks: RefCOCO~\cite{yu2016modeling}, RefCOCO+~\cite{yu2016modeling}, and RefCOCOg~\cite{mao2016generation}. \model achieves new state-of-the-art performances on most of test splits and grounding tasks.
Qualitative results and human evaluation indicate that \model is transparent and explainable.
\section{Related Work}
Visual grounding is a task that requires a system to localize a region in an image while given a natural language expression.
Different from object detection~\cite{ren2015faster}, the key for visual grounding is to utilize the linguistic information to distinguish the target from other objects, especially the objects of the same category.

To solve this problem, pioneering methods~\cite{mao2016generation, yu2016modeling, luo2017comprehension, yu2017joint} use the CNN-LSTM structure to localize the region that can generate the expression with maximum posteriori probability.
Recently, joint embedding models~\cite{hu2017modeling, yu2018mattnet, yu2018rethining} are widely used, they model the conditional probability and then localize the region with maximum probability conditioned on the expression.
Our model belongs to the second category.
However, compared with the previous works which neglect the rich linguistic structure, we step forward by taking structure information into account.
Compared to~\cite{cirik2018using} which relies on constituency parsing tree, our model applied dependency parsing tree with great parsing detail and the module assembly is learned end-to-end from scratch, while theirs is hand-crafted.

There are some works~\cite{hu2017modeling, yu2018mattnet} on using module networks in visual grounding task. However, they over-simplify the language structure and their modules are too coarse compared to ours.
Fine-grained module networks are widely used in VQA~\cite{andreas2016neural, cao2018visual, hu2017learning}. However, they rely on additional annotations to learn a sentence-to-module layout parser, which is not available in general domains. Our module layout is trained from scratch by using the Gumbel-Softmax training strategy~\cite{jang2016categorical}, which has shown empirically effective in recent works~\cite{choi2018learning, veit2018convolutional, niu2019recursive}.

\section{\model Model}
\begin{figure*}[t]
    \centering
    \includegraphics[width=\linewidth]{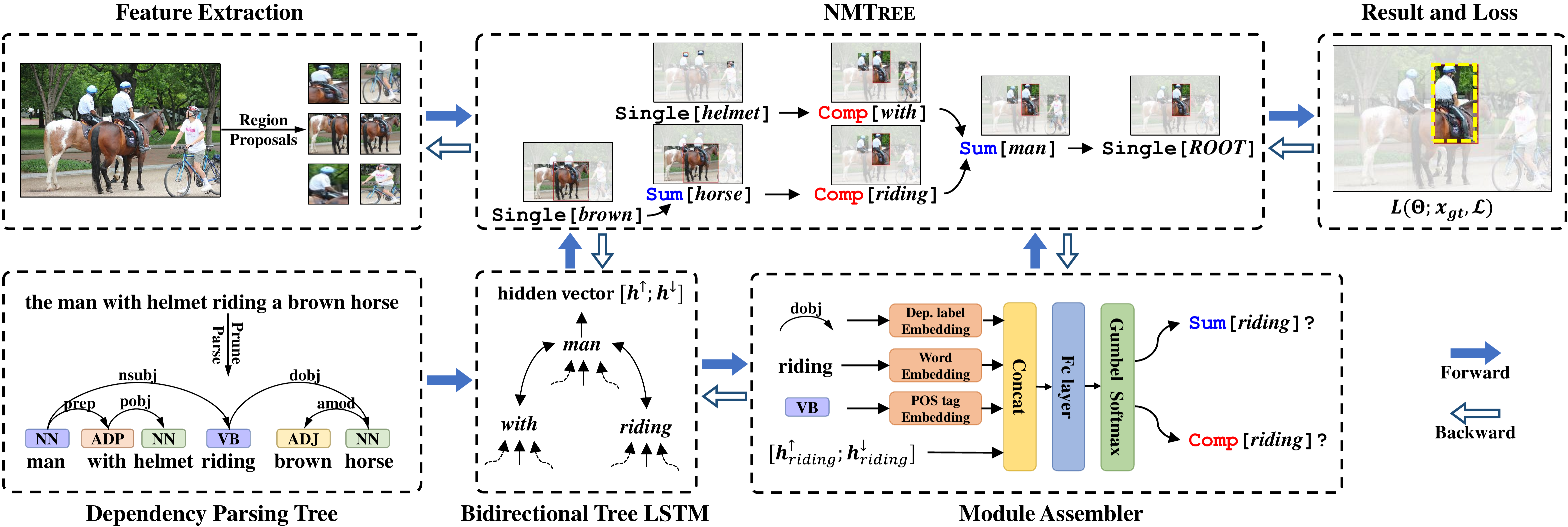}
    \caption{The overview of \model for visual grounding. Given a natural language expression as input, we first transform it into \model by Dependency Parsing Tree, Bidirectional Tree LSTM, and Module Assembler (Section~\ref{sec:3.2}). Then we ground along the tree in a bottom-up fashion (Section~\ref{sec:3.3}). The final result grounding score is the output score of the root node. We apply Gumbel-Softmax strategy to train our model (Section~\ref{sec:3.4}).}
    \label{fig:2}
    \vspace{-5pt}
\end{figure*}

In this section, we first formulate the problem of visual grounding in Section~\ref{sec:3.1}. Then, by using the walk-through example illustrated in Figure~\ref{fig:2}, we introduce how to build \model in Section~\ref{sec:3.2} and how to calculate the grounding score using \model in Section~\ref{sec:3.3}. Finally, we detail the Gumbel-Softmax training strategy in Section~\ref{sec:3.4}.

\subsection{Problem Formulation}
\label{sec:3.1}
The visual grounding task can be reduced into a retrieval problem. Formally, given an image $\mathcal{I}$, we represent it by a set of Region of Interest (RoI) features $\mathcal{I} = \{\bm{x}_1, \bm{x}_2, \cdots, \bm{x}_K\}$, where $\bm{x}_i \in \mathbb{R}^{d_x}$ and $K$ is the number of regions.
For a natural language phrase $\mathcal{L}$, we represent it by a word sequence $\mathcal{L} = \{w_1, w_2, \cdots, w_T\}$, where $T$ is the length of sentence.
Then, the task is to retrieve the target region $\bm{x}^*$ by maximizing the grounding score $S(\bm{x}_i, \mathcal{L})$ between any region and the language:
\begin{equation}
    \bm{x}^* = \mathrm{arg~max}_{\bm{x}_i\in \mathcal{I}}\, S(\bm{x}_i, \mathcal{L}).
    \label{eq:1}
\end{equation}
Therefore, the key is to define a proper $S(\cdot)$ that distinguishes the target region from others by comprehending the language composition.

The pioneering grounding models~\cite{mao2016generation, yu2016modeling} are generally based on the holistic sentence-level language representation (Figure~\ref{fig:1a}): $S(\bm{x}_i, \mathcal{L}) := S_h(\bm{x}_i, \bm{y_h})$, where $\bm{y_h}$ is a feature representation for the whole language expression and $S_h(\cdot)$ can be any similarity function between two vectors. More recently, a coarse composition~\cite{hu2017modeling} was proposed to represent the sentence as a (subject, relationship, object) triplet (Figure~\ref{fig:1b}). Thus, the score can be decomposed into a finer-grained composition: $S(\bm{x}_i, \mathcal{L}) := S_s(\bm{x}_i, \bm{y}_s) + S_r([\bm{x}_i,\bm{x}_o], \bm{y}_r) +S_o(\bm{x}_o, \bm{y}_o)$
where the subscripts $s$, $r$, and $o$ indicate the three linguistic roles: subject, relationship, and object, respectively; $\bm{x}_o$ is an estimated object region feature. However, these grounding scores over-simplify the composition of the language.
For example, as shown in Figure~\ref{fig:1b}, it is meaningful to decompose short sentences such as ``umbrella carried by girl'' into triplets, as it has a clear vision-language association for individual  ``girl'', ``umbrella'', and their relationship; but it is problematic for longer sentences that are more general with clauses, \eg, even if the ``girl in pink boots'' is identified as the object, it is still coarse and difficult for grounding.

To this end, we propose to use the Dependency Parsing Tree (DPT) as a fine-grained language decomposition, which empowers the grounding model to perform visual reasoning in great detail (Figure~\ref{fig:1c}):
\begin{equation}
S(\bm{x}_i, \mathcal{L}) := \sum\nolimits_t S_t(\bm{x}_i, \mathcal{L}_t),
\label{eq:2}
\end{equation}
where $t$ is a node in the tree, $S_t(\cdot)$ is a node-specific score function that calculates the similarity between a region and a node-specific language part $\mathcal{L}_t$. Intuitively, Eq.~\eqref{eq:2} is more human-like: accumulating the evidence (\eg, grounding score) while comprehending the language. Next, we will introduce how to implement Eq.~\eqref{eq:2}.

\subsection{Sentence to \modelend}
\label{sec:3.2}
There are three steps to transform a sentence into the proposed \modelend, as shown in the bottom three blocks of Figure~\ref{fig:2}. First, we parse the sentence into a DPT, where each word is a tree node. Then, we encode each word and its linguistic information into a hidden vector by a Bidirectional Tree LSTM. Finally, we assemble the neural modules to the tree according to node hidden vectors. 

\noindent\textbf{Dependency Parsing Tree.}
We adopt a dependency parser from Spacy toolbox\footnote{Spacy2: \url{https://spacy.io/}}. As shown in Figure~\ref{fig:2}, it structures the language into a tree, where each node is a word with its part-of-speech (POS) tag and dependency relation label of the directed edge from it to another, \eg, ``riding'' is VB (verb) and its nsubj (nominal subject) is ``man'' as NN (noun). DPT offers an in-depth comprehension of a sentence and its tree structure offers a reasoning path for visual grounding. Note that there are always unnecessary syntax elements parsed from a free-form sentence such as determiners, symbols, and punctuation. We remove these nodes and edges to reduce the computational overhead without hurting the performance.

\noindent\textbf{Bidirectional Tree LSTM.}
Once the DPT is obtained, we encode each node into a hidden vector by a bidirectional tree-structured LSTM~\cite{tai2015improved}. This bidirectional (\ie, bottom-up and top-down) propagation makes each node being aware of the information both from its children and parent. This is particularly crucial for capturing the context in a sentence. For each node $t$, we embed the word $w_t$, POS tag $p_t$, and dependency relation label $d_t$ into a concatenated embedding vector as:
\begin{equation}
    \bm{e}_t = [\mathbf{E}_w \Pi_{w_t}, \mathbf{E}_p \Pi_{p_t}, \mathbf{E}_d \Pi_{d_t}],
\label{eq:3}
\end{equation}
where $\mathbf{E}_w$, $\mathbf{E}_p$, and $\mathbf{E}_d$ are trainable embedding matrices, $\Pi_{w_t}$, $\Pi_{p_t}$, and $\Pi_{d_t}$ are one-hot encodings, for word, POS tag, and dependency relation label, respectively.

Our tree LSTM implementation \footnote{For space reasons, we leave the details in supplementary material.} is based on the Child-Sum Tree LSTM~\cite{tai2015improved}.
Taking the bottom-up direction for example, a node $t$ receives the LSTM states from its children node set $\mathcal{C}_t$ and its embedding vector $\bm{e}_t$ as input to update the state:
\begin{equation} \label{eq:4}
    \bm{c}_t^{\uparrow}, \bm{h}_t^{\uparrow} = \mathop{\textrm{TreeLSTM}}(\bm{e}_t, \{\bm{c}_{tj}^{\uparrow}\}, \{\bm{h}_{tj}^{\uparrow}\}),\quad j\in\mathcal{C}_t,
\end{equation}
where $\bm{c}_{tj}^{\uparrow},\, \bm{h}_{tj}^{\uparrow}$ denote the cell and hidden vectors of the $j$-th child of node $t$. By applying the TreeLSTM in two directions, we can obtain the final node hidden vector $\bm{h}_t$ as:
\begin{equation}
    \bm{h}_{t} = [\bm{h}_t^{\uparrow};\,\bm{h}_t^{\downarrow}],
\label{eq:5}
\end{equation}
where $\bm{h}_t^{\uparrow},\,\bm{h}_t^{\downarrow}\in\mathbb{R}^{d_h}$ denote the hidden vectors encoded in the bottom-up  and top-down directions, respectively. We initialize all leaf nodes with zero hidden and cell states. The bottom-up and top-down Tree LSTMs have their independent trainable parameters.

\noindent\textbf{Module Assembler.}
Given the node representation $\bm{e}_t$ and the above obtained node hidden vector $\bm{h}_t$, we can feed them into a module assembler to determine which module should be assembled to node $t$. As we will detail in Section~\ref{sec:3.3}, we have three modules, \textit{i.e.}, \texttt{Single}, \texttt{Sum}, and \texttt{Comp}. Since the \texttt{Single} is always assembled on leaves and the root, the assembler only need to choose between \texttt{Sum} and \texttt{Comp} as:
\begin{equation}
    \!\!\texttt{Sum}~\textrm{or}~\texttt{Comp} \leftarrow \mathrm{arg~max}~\textrm{softmax}\left(\textrm{fc}([\bm{e}_t, \bm{h}_t])\right),
\label{eq:6}
\end{equation}
where $\textrm{fc}$ is a fully connected layer that maps the input features into a 2-d values, indicating the relative scores for \texttt{Sum} and \texttt{Comp}, respectively.
Due to the discrete and non-differentiable nature of $\mathrm{arg~max}$, we use the Gumbel-Softmax~\cite{jang2016categorical} strategy for training (Section~\ref{sec:3.4}).

\begin{figure}[t]
\centering
\subfigure[\texttt{Sum} Module Node]{
  \centering
  \fbox{\includegraphics[width=.45\linewidth]{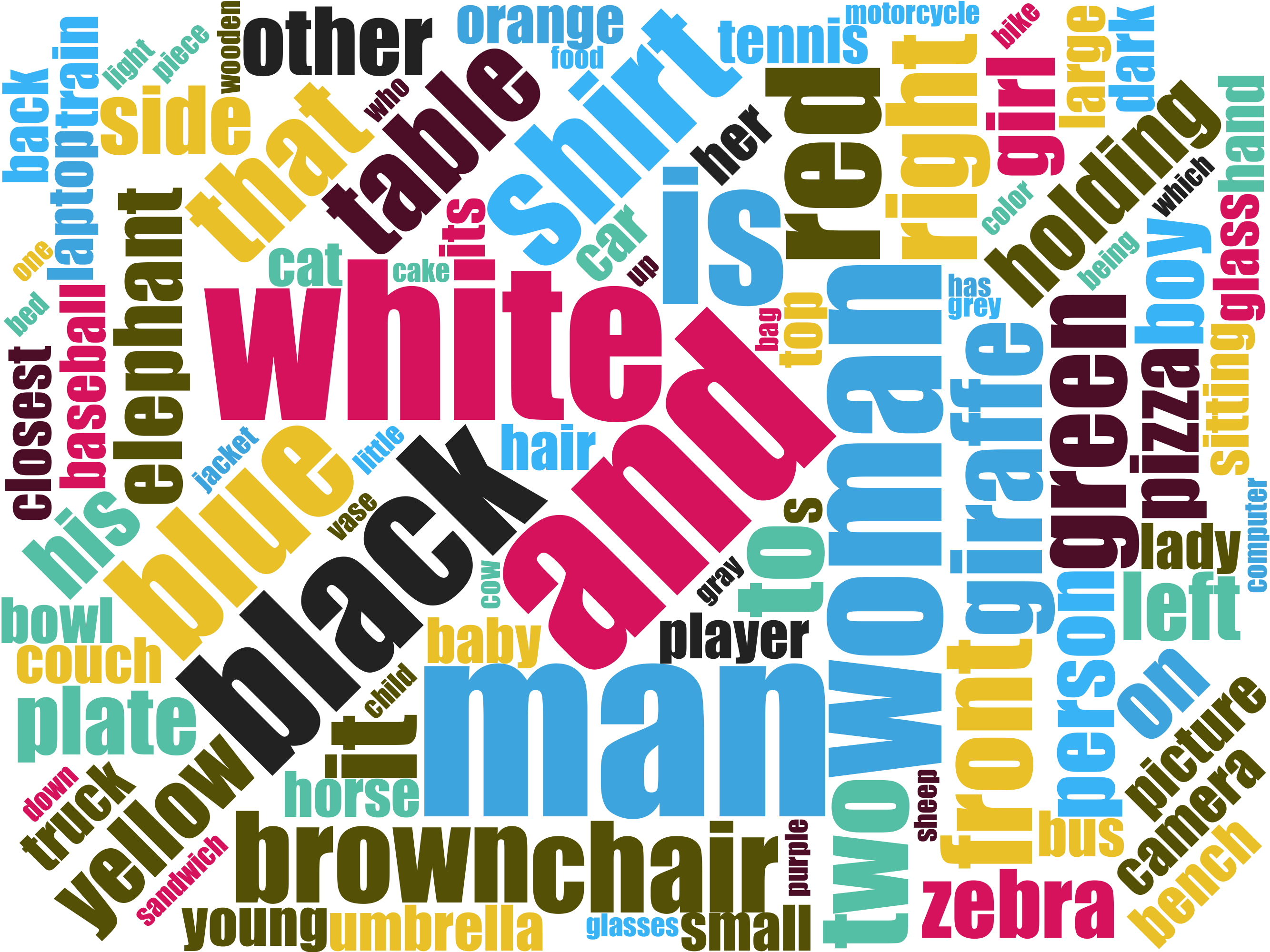}}
  \label{fig:3a}
}%
\subfigure[\texttt{Comp} Module Node]{
  \centering
  \fbox{\includegraphics[width=.45\linewidth]{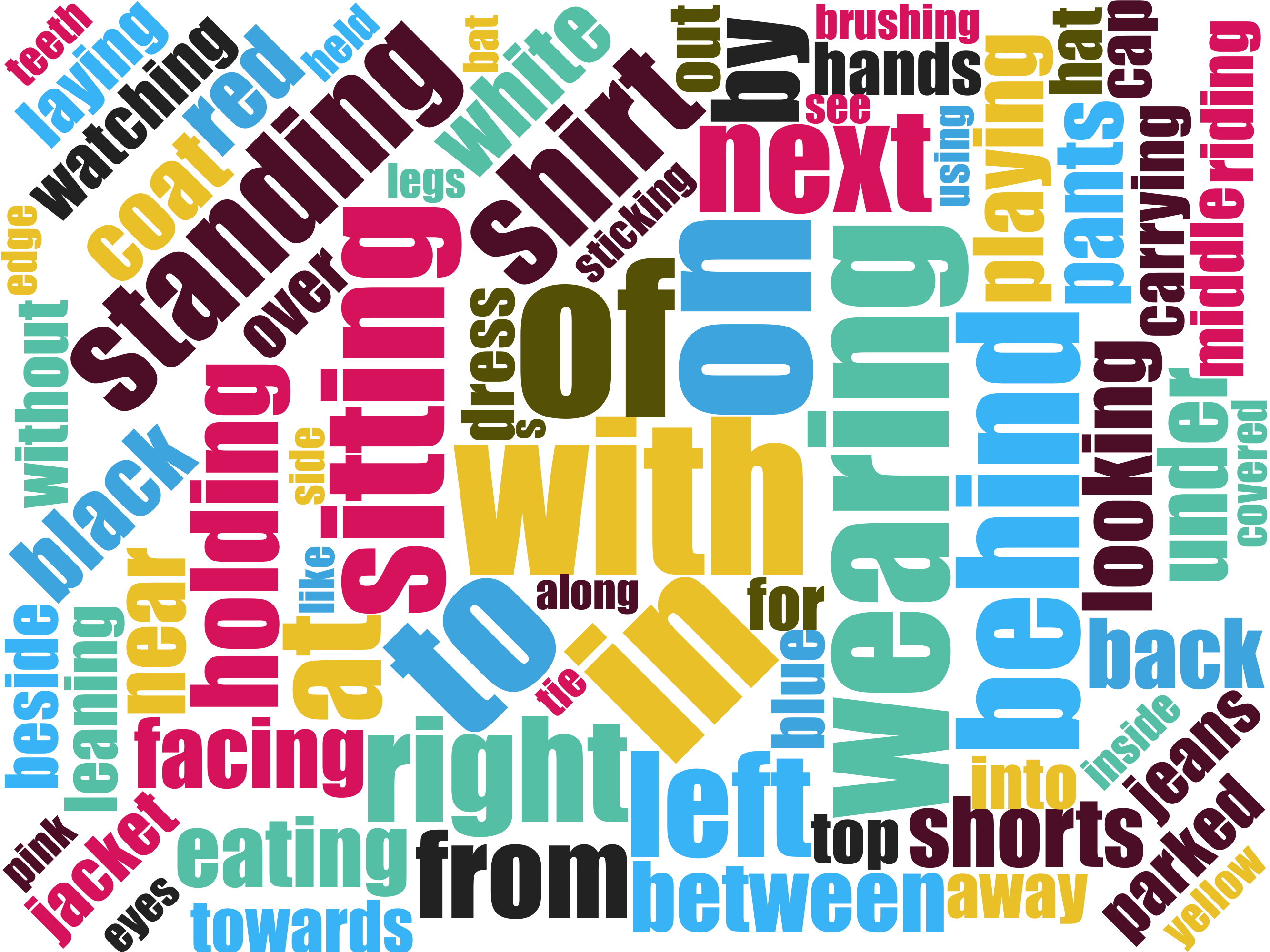}}
  \label{fig:3b}
}
\caption{Word cloud visualizations of what nodes are likely to be assembled as \texttt{Sum} or \texttt{Comp}. We can find that the \texttt{Sum} module nodes are likely to be visible concepts while the \texttt{Comp} module nodes are likely to be relationship concepts.}
\vspace{-10pt}
\label{fig:3}
\end{figure}

It is worth noting that the assembler is not purely linguistic even though Eq.~\eqref{eq:6} is based on DPT node features. In fact, thanks to the back-propagation training algorithm, visual cues will be eventually incorporated into the parameters of Eq.~\eqref{eq:6}.
Figure~\ref{fig:3} illustrates which type of words is likely to be assembled by each module. We can find that the \texttt{Sum} module has more visible words (\eg, adjectives and nouns), and the \texttt{Comp} module has more words describing relations (\eg, verbs and prepositions).
This reveals the explainable potential of \modelend.
Finally, by the above three steps, we get the \model that each node assembled. Next, we will elaborate the three types of modules.

\subsection{\model Modules}
\label{sec:3.3}
Given the above assembled \modelend, we can implement the tree grounding score proposed in Eq.~\eqref{eq:2} by accumulating the scores in a bottom-up fashion.
There are three types of modules used in \modelend, \ie, \texttt{Single}, \texttt{Sum} and \texttt{Comp}. Each module at node $t$ updates the grounding score $\bm{s}_t = [s_t^1, \cdots, s_t^K]$ for all the $K$ regions in the image $\mathcal{I}$ and outputs to its parent.
In the following, we will first introduce language representation and common functions used in the modules, and then detail each module.

\noindent \textbf{Language Representation.} 
For node $t$, we have two language representations: $\bm{y}^{s}_{t}$ is used to associate with a single visual feature and $\bm{y}^{p}_{t}$ is used to associate with a pairwise visual feature.
We denote the node set of node $t$ as $\mathcal{N}_{t}$, which contains itself and all nodes rooted from $t$.
Therefore, the language representation can be calculated by the weighted sum of node embedding vectors from $\mathcal{N}_t$:
\begin{equation}
    \bm{y}^{s}_{t} = \sum\nolimits_{i \in\mathcal{N}_t} \alpha^s_{i} \bm{e}_{i}, \quad
    \bm{y}^{p}_{t} = \sum\nolimits_{i\in\mathcal{N}_t} \alpha^p_{i} \bm{e}_{i},
    \label{eq:7}
\end{equation}
where $\alpha$ are the node-level attention weights that calculated from the corresponding node hidden vectors: $\alpha_{i}~=~\mathrm{softmax}(\mathrm{fc}(\bm{h}_i))$.
Note that $\alpha^s_{i}$ and $\alpha^p_{i}$ have independent fc parameters. It is worth noting that these weighted average word embeddings of the node set reduce the negative impact caused by DPT parsing errors~\cite{hu2017modeling}.

\noindent\textbf{Score Functions.} There are two types of score functions used in our modules, denoted by the single score function $S_s$ and pairwise score function $S_p$, where $S_s$ measures the similarity between a single region $\bm{x}$ and a language representation $\bm{y}$, and $S_p$ indicates how likely pair-wise regions match with relationships.
Formally we define them as:
\begin{equation}
    S_s(\bm{x}, \bm{y}) = \textrm{fc}(\textrm{L2norm}(\textrm{fc}(\bm{x}) \odot \bm{y})),
    \label{eq:8}
\end{equation}
\begin{equation}
    S_p(\bm{x}_1, \bm{x}_2, \bm{y}) = \textrm{fc}(\textrm{L2norm}(\textrm{fc}([\bm{x}_1; \bm{x}_2]) \odot \bm{y})),
    \label{eq:9}
\end{equation}
where $[;]$ is a concatenation operation, $\odot$ is element-wise multiplication, and $\textrm{L2norm}$ is used to normalize vectors.

\noindent\textbf{\texttt{Single} Module}. It is assembled at leaves and the root.
Its job is to 1) calculate a single score for each region and the current language feature by Eq.~\eqref{eq:8}, 2) add this new score to the scores collected from children,  and then 3) pass the sum to its parent:
\begin{equation}
\begin{split}
\textbf{Input:} \quad &\{\bm{s}_{tj}\}, \quad j \in \mathcal{C}_t\\
\textbf{Output:} \quad & s_t^i \leftarrow S_s(\bm{x}_i ,\bm{y}_t^s) + \sum\nolimits_{j} s_{tj}^i,  i \in [1, K]
\end{split}
\label{eq:10}
\end{equation}
Note that for leaves, $\mathcal{C}_{t} = \phi$ as they have no children.
As illustrated in Figure~\ref{fig:2}, its design motivation is to initiate the bottom-up grounding process by the most elementary words and finalize the grounding by passing the accumulated scores to ROOT.

\noindent\textbf{\texttt{Sum} Module}. It plays a transitional role during the reasoning process. It simply sums up the scores passed from its children and then passes the sum to its parent:
\begin{equation}
\begin{split}
\textbf{Input:} \quad & \{\bm{s}_{tj}\}, \quad j \in \mathcal{C}_t\\
\textbf{Output:} \quad & \bm{s}_t \leftarrow \sum\nolimits_{j}\bm{s}_{tj}
\end{split}
\label{eq:11}
\end{equation}
Note that this module has no parameters hence it significantly reduces the complexity of our model.
As illustrated in Figure~\ref{fig:2}, intuitively, it transits the easy-to-localize words (cf. Figure~\ref{fig:3a}) such as ``horse'' and ``man'' to help the subsequent composite grounding.

\noindent\textbf{\texttt{Comp} Module.} This is the core module for composite visual reasoning. As shown in Figure~\ref{fig:3b}, it is likely to be the relationship that connects two language constitutions. It first computes an ``average region'' visual feature that is grounded by the single scores:
\begin{equation}
\!\!\beta_i = \textrm{softmax}\!\left(\!S_s(\bm{x}_i, \bm{y}_t^s)\! +\!\! \sum\nolimits_{j} \!\!s_{tj}^{i}\right),~\bar{\bm{x}}\! = \!\sum\nolimits_{i} \beta_i \bm{x}_i.
\label{eq:12}
\end{equation}
In particular, $\bar{\bm{x}}$ can be considered as the contextual region~\cite{zhang2018grounding} that supports the target region score, \eg, ``what is riding the horse'' in Figure~\ref{fig:2}. Therefore, this module outputs the target region score to its parent:
\begin{equation}
\begin{split}
\textbf{Input:} \quad & \{\bm{s}_{tj}\}, \quad j \in \mathcal{C}_t\\
\textbf{Output:} \quad & s_t^i \leftarrow S_p(\bm{x}_i, \bar{\bm{x}} ,\bm{y}_t^p).
\end{split}
\label{eq:13}
\end{equation}
Recall that $\mathbf{y}^p_t$ is pairwise language feature that represents the relationship words. 

\begin{figure}
    \centering
    \includegraphics[width=0.95\linewidth]{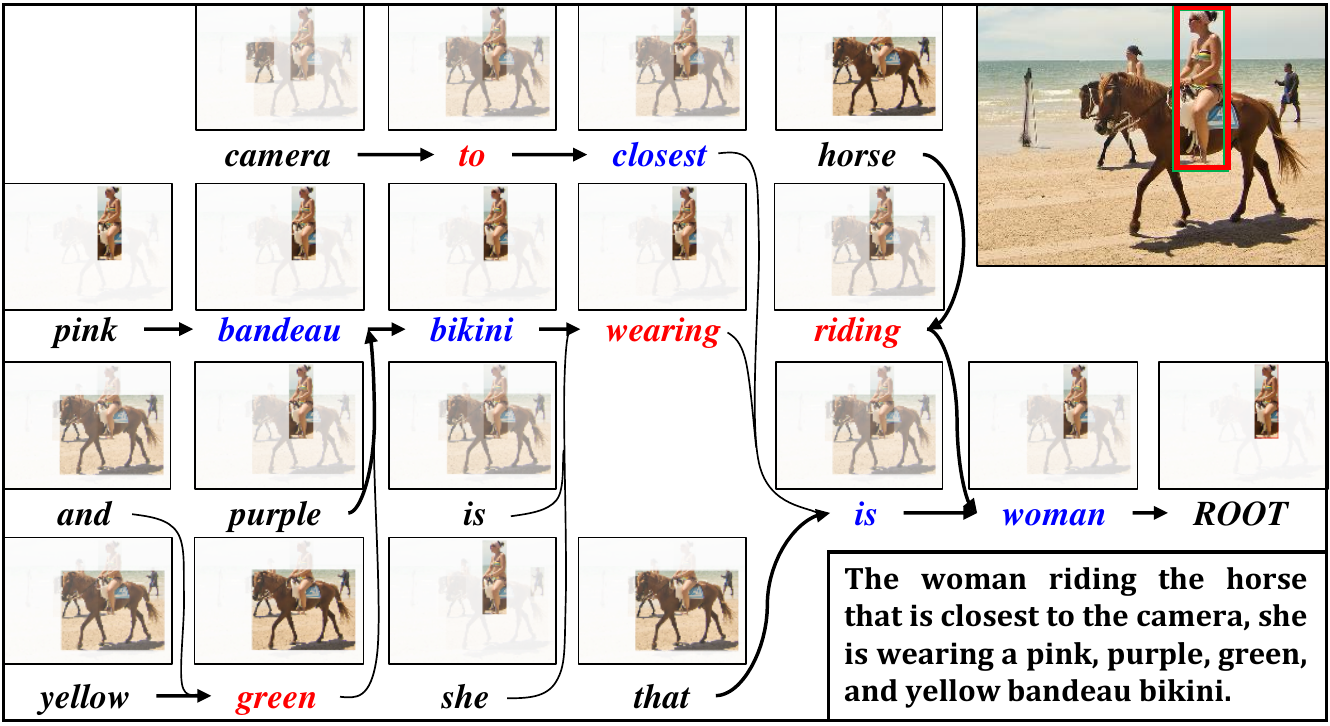}
    \caption{A very long sentence example that illustrates the explainability of the neural modules in \modelend. Black words:\texttt{Single}, blue words: \texttt{Sum}, red words: \texttt{Comp}.}
    \vspace{-10pt}
    \label{fig:4}
\end{figure}

By reasoning along the assembled \model in bottom up fashion, we can obtain the overall accumulated grounding score in Eq.~\eqref{eq:2} at tree root.
Moreover, thanks to the score output at each node, \model is transparent as the scores can be visualized as attention maps to investigate the grounding process.
Figure~\ref{fig:4} illustrates an extreme example with a very long expression with 22 tokens. However, by using the neural modules in \model, it still works well and reasons with explainable intermediate process. Next, we will discuss how to train \modelend.

\subsection{\model Training}
\label{sec:3.4}
In contrast to previous neural module networks~\cite{andreas2016neural, hu2017learning}, \model does not require any additional annotations and is end-to-end trainable. Suppose $\bm{x}_{gt}$ is the ground-truth region, the objective is to minimize the cross-entropy loss:
\begin{equation}
    L(\Theta;\bm{x}_{gt}, \mathcal{L}) = - \log\textrm{softmax}(S(\bm{x}_{gt}, \mathcal{L};\Theta)),
    \label{eq:14}
\end{equation}
where $\Theta$ is the trainable parameter set and softmax is across all $K$ regions in an image. 

Recall that the assembling process in Eq.~\eqref{eq:6} is discrete and blocks the end-to-end training. Therefore, we utilize the Gumbel-Softmax strategy~\cite{gumbel1954statistical} that is shown effective in recent works~\cite{choi2018learning, veit2018convolutional} on architecture search. For more details, please refer to their papers. Here, we only introduce how to apply the Gumbel-Softmax for \model training.

\noindent\textbf{Forward}. We add Gumbel distribution as a noise into the relative scores (\ie $\textrm{fc}([\bm{e}_t, \bm{h}_t])$) of each module. It introduces stochasticity for the module assembling exploration. Specifically, we parameterize the assembler decision as a 2-d one-hot vector $\bm{z}$, where the index of non-zero entry indicates the decision:
\begin{equation}
    \bm{z} = \mathtt{one\_hot}(\mathrm{arg~max}(\log(\textrm{fc}([\bm{e}_t, \bm{h}_t]))+G)),
    \label{eq:15}
\end{equation}
where $G$ is the noise drawn from i.i.d. Gumbel(0, 1)\footnote{The Gumbel~(0, 1) distribution is sampled by $G=-\log(-\log(U))$ where $U \sim \mathrm{Uniform}(0,1)$.}. Note that, in inference phrase, the $G$ will be discarded.

\noindent\textbf{Backward.} We take a continuous approximation that relaxes $\bm{z}$ to $\widetilde{\bm{z}}$ by replacing argmax with softmax, formally:
\begin{equation}
    \widetilde{\bm{z}} = \textrm{softmax}((\log(\textrm{fc}([\bm{e}_t, \bm{h}_t]))+G)/\tau),
    \label{eq:16}
\end{equation}
where $G$ is the same sample drawn in the forward pass (\ie, we reuse the noise samples). $\tau$ is a temperature parameter that the softmax function approaches to argmax while $\tau \rightarrow 0$ and approaches to uniform while $\tau \rightarrow \infty$.
Although there are discrepancies between the forward and backward pass, we empirically observe that the Gumbel-Softmax strategy performs well in our experiments.

\section{Experiments}

\begin{table*}[ht]
\footnotesize
\begin{center}
\resizebox{2.0\columnwidth}{!}{%
\begin{tabular}{| l | c | c | c | c | c | c | c | c || c | c | c | c | c | c | c | c |}
\hline
& \multicolumn{3}{c}{RefCOCO} & \multicolumn{3}{|c|}{RefCOCO+} & \multicolumn{2}{|c||}{RefCOCOg} & \multicolumn{3}{c}{RefCOCO(det)} & \multicolumn{3}{|c|}{RefCOCO+(det)} & \multicolumn{2}{|c|}{RefCOCOg(det)}\\
\cline{1-17}
& val & testA & testB & val & testA & testB & val & test & val & testA & testB & val & testA & testB & val & test\\
\hline\hline
Chain & 82.43 & 82.21 & 82.16 & 68.27 & 70.83 & 62.41 & 73.84 & 74.15 & 74.81 & 79.19 & 68.34 & 63.08 & 68.84 & 53.53 & 61.72 & 61.95\\
\model w/o \texttt{Comp} & 83.65 & 83.59 & 83.04 & 70.76 & 73.07 & 65.19 & 75.98 & 76.20 & 75.10 & 79.38 & 68.60 & 64.85 & 70.43 & 55.00 & 63.07 & 63.40\\
\model w/o \texttt{Sum} & 83.79 & 83.81 & 83.67 & 70.83 & 73.72 & 65.83 & 76.11 & 76.09 & 75.49 & 79.84 & 69.11 & 65.29 & 70.85 & 55.99 & 63.60 & 64.06 \\
\model w/ Rule & 84.46 & 84.59 & 84.26 & 71.48 & 74.76 & 66.95 & 77.82 & 77.70 & 75.51 & 80.61 & 69.23 & 65.23 & 70.94 & 56.96 & 64.69 & 65.53 \\
\model & \textbf{85.65} & \textbf{85.63} & \textbf{85.08} & \textbf{72.84} & \textbf{75.74} & \textbf{67.62} & \textbf{78.57} & \textbf{78.21} & \textbf{76.41} & \textbf{81.21} & \textbf{70.09} & \textbf{66.46} & \textbf{72.02} & \textbf{57.52} & \textbf{65.87} & \textbf{66.44}\\
\hline
\end{tabular}
}
\end{center}
\vspace{-2pt}
\caption{Top-1 Accuracy\% of ablation models on the three datasets.}
\vspace{-5pt}
\label{table:1}
\end{table*}

\subsection{Datasets}
We conducted our experiments on three datasets that are collected from MS-COCO~\cite{lin2014microsoft} images.
\textbf{RefCOCO}~\cite{yu2016modeling} contains 142,210 referring expressions
for 19,994 images. An interactive game~\cite{kazemzadeh2014referitgame} is used during the expression collection.
All expression-referent pairs are split into train, validation, testA, and testB.
TestA contains the images with multiple people and testB contains the images with multiple objects.
\textbf{RefCOCO+}~\cite{yu2016modeling} contains 141,564 referring expressions for 49,856 objects in 19,992 images.
It is collected with the same interactive game as RefCOCO and is split into train, validation, testA, and testB, respectively.
The difference from RefCOCO is that RefCOCO+ only allows expression described by appearance but no locations.
\textbf{RefCOCOg}~\cite{mao2016generation} contains 95,010 referring expressions for 49,822 objects in 25,799 images.
It is collected in a non-interactive way and contains longer expressions described by both appearance and locations.
It has two types of data partitions.
The first partition~\cite{mao2016generation} divides dataset
into train and validation (val$^*$) sets.
The second partition~\cite{nagaraja2016modeling} divides images into train, validation (val) and test sets.

\subsection{Implementation Details and Metrics}
\noindent\textbf{Language Settings.}
We built specific vocabularies for the three datasets with words, POS tags, and dependency labels appeared more than once in datasets.
Note that to obtain accurate parsing results, we did not trim the length of expressions.
We used pre-trained GloVe~\cite{pennington2014glove} to initialize word vectors.
For dependency label vectors and POS tag vectors, we trained them from scratch with random initialization.
We set the embedding sizes to 300, 50, 50 for words, POS tags, and dependency labels, respectively.

\noindent\textbf{Visual Representations.}
To represent RoI features of an image, we concatenated object features and location features extracted from MAttNet~\cite{yu2018mattnet}, which is based on Faster RCNN~\cite{ren2015faster} with ResNet-101~\cite{he2016deep} as the backbone and trained with attribute heads.
We employed Mask RCNN~\cite{he2017mask} for object segmentation.
The visual feature dimension $d_x$ was set to 3,072.
For fair comparison, we also used VGG-16~\cite{simonyan2014very} as the backbone and $d_x$ was set to 5,120.

\noindent\textbf{Parameter Settings.}
We optimized our model with Adam optimizer~\cite{kingma2014adam} up to 40 epochs. The learning rate was initialized to 1e-3 and shrunk by 0.9 every 10 epochs. We set 128 images to the mini-batch size. 
The LSTM hidden size $d_h$ was set to 1,024, the hidden size of the attention in language representation was set to 1,024.
The temperature $\tau$ of Gumbel-Softmax~\cite{jang2016categorical} was set to 1.0.

\noindent\textbf{Evaluation Metrics.}
For detection task, we calculated the Intersection-over-Union (IoU) between the detected bounding box and the ground-truth one, and treated the one with IoU at least 0.5 as correct. We used the Top-1 accuracy as the metric, which is the fraction of the correctly grounded test expressions.
For segmentation task, we used Pr@0.5 (the percentage of expressions where IoU at least 0.5) and overall IoU as metrics.

\begin{table*}
\begin{center}
\resizebox{2.0\columnwidth}{!}{%
\begin{tabular}{| l | c | c | c | c | c | c | c | c | c || c | c | c | c | c | c | c | c | c |}
\hline
 & \multicolumn{3}{c}{RefCOCO} & \multicolumn{3}{|c|}{RefCOCO+} & \multicolumn{3}{|c||}{RefCOCOg} & \multicolumn{3}{c}{RefCOCO (det)} & \multicolumn{3}{|c|}{RefCOCO+ (det)} & \multicolumn{3}{|c|}{RefCOCOg (det)}\\
\cline{1-19}
 & val & testA & testB & val & testA & testB & val* & val & test & val & testA & testB & val & testA & testB & val* & val & test\\
\hline\hline
MMI~\cite{mao2016generation}       & - & 63.15 & 64.21 & - & 48.73 & 42.13 & 62.14 & - & - & - & 64.90 & 54.51 & - & 54.03 & 42.81 & 45.85 & - & - \\
Attribute~\cite{liu2017referring} & - & 78.85 & 78.07 & - & 61.47 & 57.22 & 69.83 & - & - & - & 72.08 & 57.29 & - & 57.97 & 46.20 & 52.35 & - & - \\
Listener$^{\dagger}$~\cite{yu2017joint}& 78.36 & 77.97 & 79.86 & 61.33 & 63.10 & 58.19 & 72.02 & 71.32 & 71.72 & 68.95 & 72.95 & 62.98 & 54.89 & 59.61 & 48.44 & 58.32 & 59.33 & 59.21 \\
\hline
NegBag~\cite{nagaraja2016modeling} & 76.90 & 75.60 & 78.80 & - & - & - & - & - & 68.40 & 57.30 & 58.60 & 56.40 & - & - & - & 39.50 & - & 49.50 \\
\hline
CMN~\cite{hu2017modeling}  & - & 75.94 & 79.57 & - & 59.29 & 59.34 & 69.30 & - & - & - & 71.03 & 65.77 & - & 54.32 & 47.76 & 57.47 & - & - \\
VC~\cite{zhang2018grounding} & - & 78.98 & 82.39 & - & 62.56 & 62.90 & 73.98 & - & - & - & 73.33 & 67.44 & - & 58.40 & 53.18 & 62.30 & - & - \\
AccumAttn~\cite{deng2018visual} & 81.27 & 81.17 & 80.01 & 65.56 & 68.76 & 60.63 & 73.18 & - & - & - & - & - & - & - & - & - & - & - \\
MAttN$^{\ddagger}$~\cite{yu2018mattnet} & 85.65 & 85.26 & 84.57 & 71.01 & 75.13 & 66.17 & - & 78.10 & 78.12 & 76.40 & 80.43 & 69.28 & 64.93 & 70.26 & 56.00 & - & \textbf{66.67} & \textbf{67.01} \\
\hline
GroundNet~\cite{cirik2018using}  & - & - & - & - & - & - & 68.90 & - & - & - & - & - & - & - & - & - & - & - \\
parser+CMN~\cite{hu2017modeling}  & - & - & - & - & - & - & 53.50 & - & - & - & - & - & - & - & - & - & - & - \\
parser+MAttN$^{\ddagger}$~\cite{yu2018mattnet} & 80.20 & 79.10 & 81.22 & 66.08 & 68.30 & 62.94 & - & 73.82 & 73.72 & - & - & - & - & - & - & - & - & - \\
\hline
\model & 80.39 & 78.86 & 81.90 & 63.31 & 63.59 & 63.04 & 73.71 & 73.39 & 72.29 & 71.65 & 74.81 & 67.34 & 58.00 & 61.09 & 53.45 & 61.20 & 61.01 & 61.46\\
\model$^{\ddagger}$ & \textbf{85.65} & \textbf{85.63} & \textbf{85.08} & \textbf{72.84} & \textbf{75.74} & \textbf{67.62} & \textbf{78.03} & \textbf{78.57} & \textbf{78.21} & \textbf{76.41} & \textbf{81.21} & \textbf{70.09} & \textbf{66.46} & \textbf{72.02} & \textbf{57.52} & \textbf{64.62} & 65.87 & 66.44 \\
\hline
\end{tabular}}
\end{center}
% \vspace{-10pt}
\caption{Top-1 Accuracy\% of various grounding models on the three datasets. For fair comparison, we use $^{\dagger}$ to indicate that this model uses res101 features for detected experiments. $^{\ddagger}$ indicates that this model uses res101 features for both ground-truth and detected experiments. None-superscript indicates that this model uses vgg16 features. Parser+ indicates that the model used an external parser.}
% \vspace{-10pt}
\label{table:2}
\end{table*}

\begin{table}
\centering
\resizebox{1.0\columnwidth}{!}{
\begin{tabular}{|c|l|l|l|l|l|l|l|l|l|}
\hline
\multicolumn{2}{|c|}{} & \multicolumn{3}{c|}{RefCOCO} & \multicolumn{3}{c|}{RefCOCO+} & \multicolumn{2}{c|}{RefCOCOg} \\ \hline
\multicolumn{2}{|c|}{} & \multicolumn{1}{c|}{val} & \multicolumn{1}{c|}{testA} & \multicolumn{1}{c|}{testB} & \multicolumn{1}{c|}{val} & \multicolumn{1}{c|}{testA} & \multicolumn{1}{c|}{testB} & \multicolumn{1}{c|}{val} & \multicolumn{1}{c|}{test} \\ \hline\hline
\parbox[t]{2mm}{\multirow{3}{*}{\rotatebox[origin=c]{90}{\textbf{Pr@0.5}}}} & MAttNet~\cite{yu2018mattnet} & \textbf{75.16} & 79.55 & 68.87 & 64.11 & 70.12 & 54.82 & \textbf{64.48} & \textbf{65.60} \\
& Chain & 73.36 & 77.55 & 67.30 & 61.60 & 67.15 & 52.24 & 59.64 & 60.29 \\
& \model & 74.71 & \textbf{79.71} & \textbf{68.93} & \textbf{65.06} & \textbf{70.24} & \textbf{56.15} & 63.77 & 64.63 \\ \hline
\parbox[t]{2mm}{\multirow{3}{*}{\rotatebox[origin=c]{90}{\textbf{IoU}}}} & MAttNet~\cite{yu2018mattnet} & 56.51 & 62.37 & 51.70 & 46.67 & 52.39 & 40.08 & \textbf{47.64} & \textbf{48.61} \\
& Chain & 55.29 & 60.99 & 51.36 & 44.74 & 49.83 & 38.50 & 42.55 & 43.99 \\
& \model & \textbf{56.59} & \textbf{63.02} & \textbf{52.06} & \textbf{47.40} & \textbf{53.01} & \textbf{41.56} & 46.59 & 47.88 \\ \hline
\end{tabular}
}
\vspace{5pt}
\caption{Segmentation performance(\%) on the three datasets comparing with state-of-the-arts.}
\vspace{-10pt}
\label{table:3}
\end{table}

\begin{figure*}[htb]
    \centering
    \includegraphics[width=\linewidth]{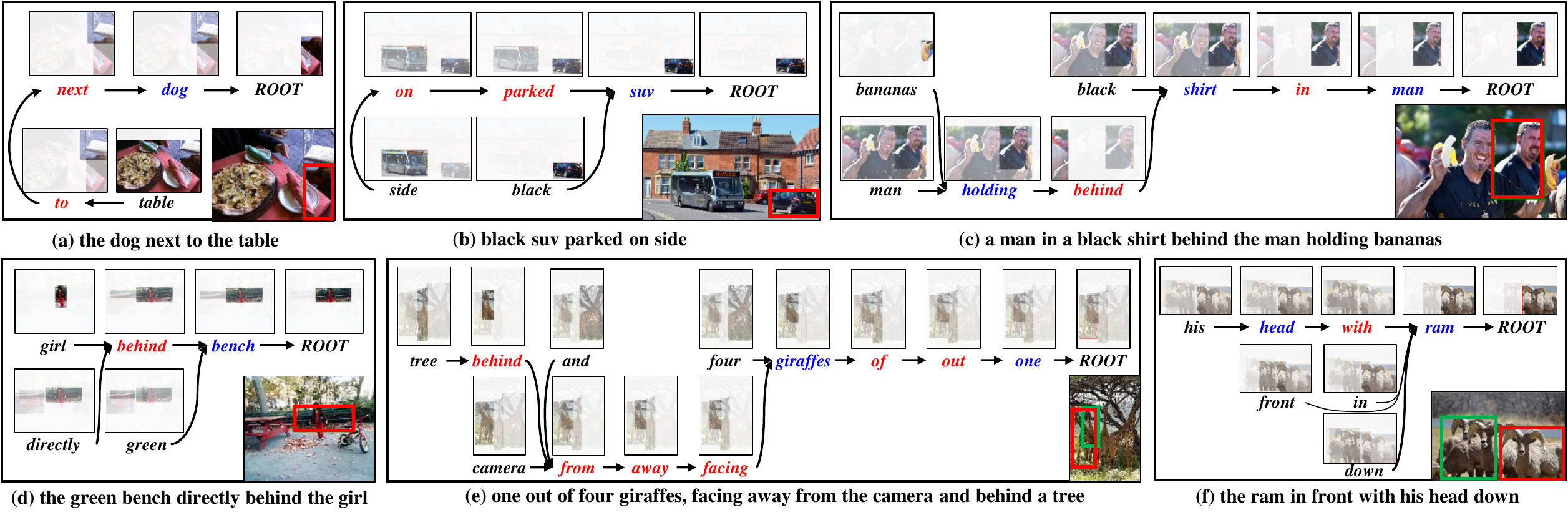}
    % \vspace{-5pt}
    \caption{Qualitative results on RefCOCOg. Words in different colors indicate corresponding modules: black for \texttt{Single}, red for \texttt{Comp}, and blue for \texttt{Sum}. The bottom right corner is the original image with a green bounding box as ground-truth and a red bounding box as our result. We further give two failure examples (e) and (f) for comparison, and our model consistently provides explainable reasoning process.}
    \vspace{-5pt}
    \label{fig:5}
\end{figure*}

\definecolor{olivegreen}{RGB}{0,128,0}
\begin{figure}[htb]
    \centering
    \includegraphics[width=\linewidth]{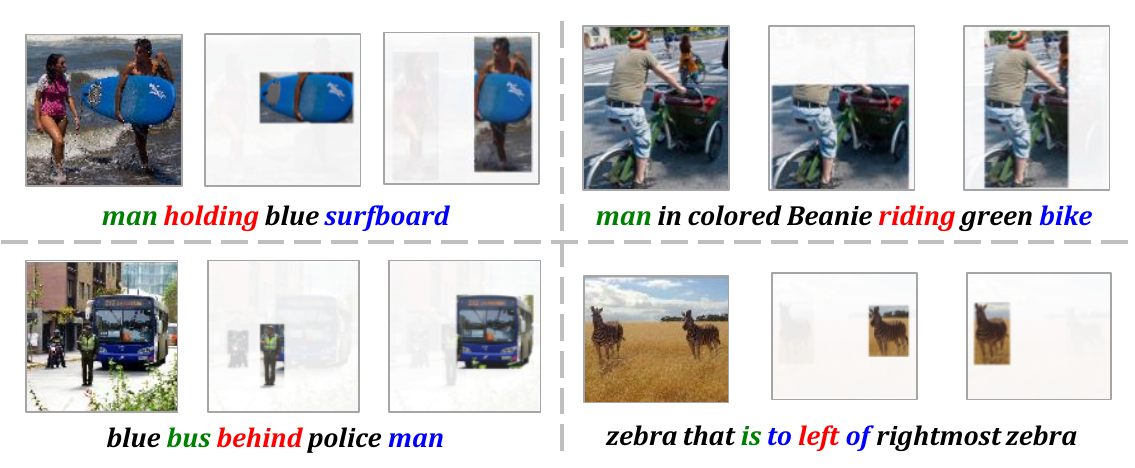}
    % \vspace{-5pt}
    \caption{The compositional reasoning inside \texttt{Comp}. Each example contains the original image (left), the contextual attention map of $\bm{\bar{x}}$ in Eq.~\eqref{eq:12} (middle) and the output attention map (right). We represent partial tree structure by colors: red for the current node, blue for children and green for parent.}
    \vspace{-10pt}
    \label{fig:6}
\end{figure}

\begin{figure}[t]
    \centering
    \includegraphics[width=0.85\linewidth]{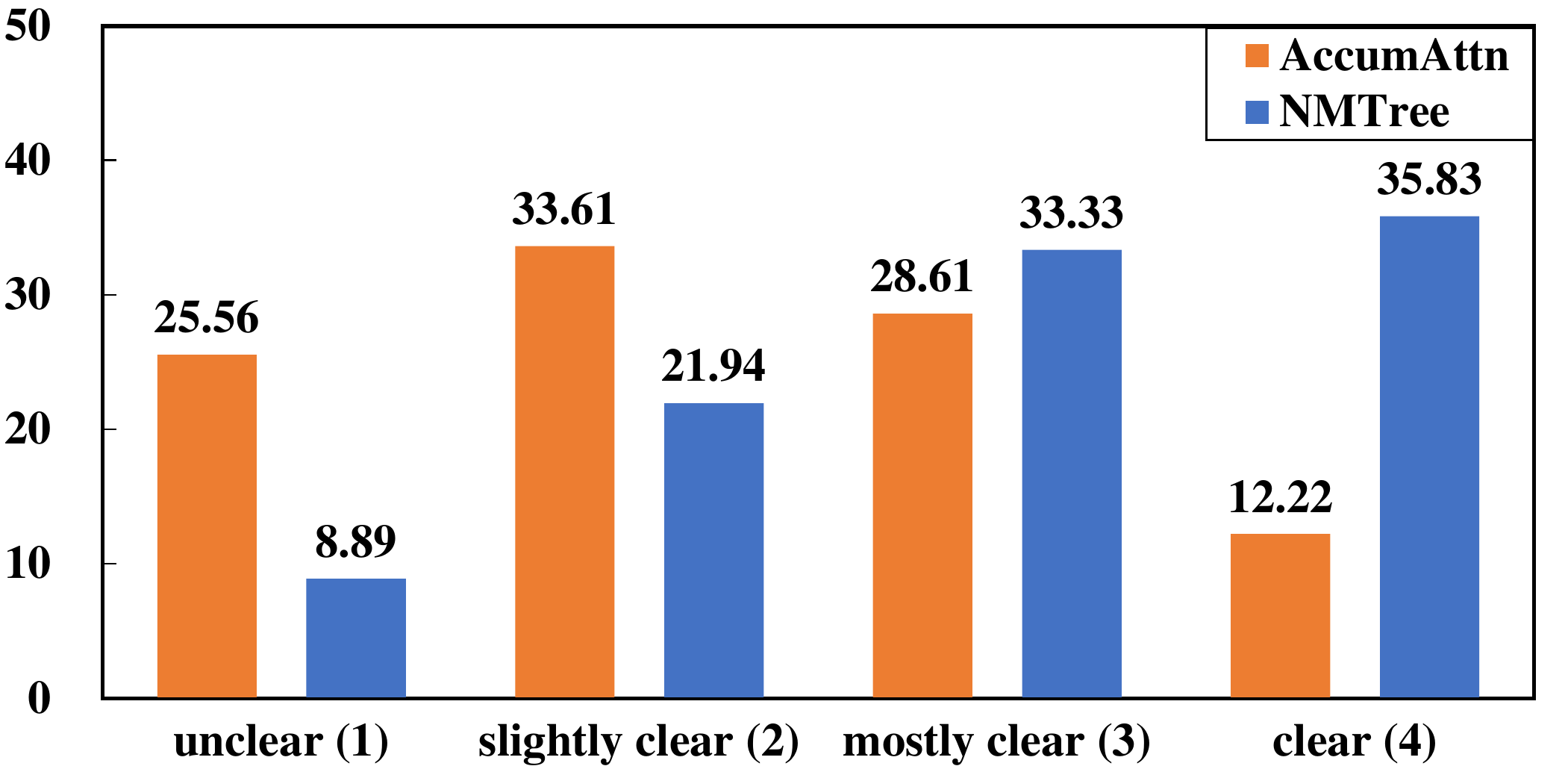}
    \caption{The percentage of each choice. The average scores of \model and AccumAttn~\cite{deng2018visual} are \textbf{2.96} and \textbf{2.28}. The results indicate that our model is more explainable to humans.}
    \vspace{-10pt}
    \label{fig:7}
\end{figure}

\subsection{Ablation Studies}
\label{sec:4.3}
\noindent\textbf{Settings}.
We conducted extensive ablation studies to reveal the internal mechanism of \modelend. The ablations and their motivations are detailed as follows.
\textbf{Chain}: it ignores the structure information of the language. 
Specifically, we represent a natural language expression as the weighted average of each word embedding, where the weights are calculated by soft attention on bi-LSTM hidden vectors of each word. The final grounding score is calculated by single score function between each region and the language representation.
\textbf{\model w/o \texttt{Comp}}: it is the \model without the \texttt{Comp} module, forcing all internal nodes as \texttt{Sum} module. \textbf{\model w/o \texttt{Sum}}: it is the \model without the \texttt{Sum} module, forcing all internal nodes as \texttt{Comp} module. \textbf{\model w/ Rule}: it assembles modules by a hand-crafted rule. Instead of deciding which module should be assembled to each node by computing the relative score, we designed a fixed linguistic rule to make a discrete and non-trainable decisions. The rule is: set the internal nodes whose dependency relation label is `acl' (\ie, adjectival clause) or `prep' (\ie, prepositional modifier) as \texttt{Comp} module, and the others as \texttt{Sum} module.

\noindent\textbf{Results.}
Table~\ref{table:1} shows the grounding accuracies of the ablation methods on the three benchmarks. We can have the following observations:
1) On all datasets, \model outperforms Chain even if we removed one module or used the hand-crafted rule. This is because the tree structure contains more linguistic information and more suitable for reasoning. Meanwhile, it also demonstrates that our proposed fine-grained composition is better than the holistic Chain.
2) When we removed one module, \ie, \model w/o \texttt{Comp} and \model w/o \texttt{Sum}, they are worse than the full \modelend. It demonstrates the necessity of the \texttt{Sum} and \texttt{Comp}. Note that removing any modules will also hurt the explainability of models.
3) \model w/o \texttt{Comp} and \model w/o \texttt{Sum} are comparable but \model w/o \texttt{Sum} is slightly better. This is because the \texttt{Comp} module is more complex and thus resulting in overfitting.
4) \model outperforms \model w/ Rule. It demonstrates that \model can automatically find which nodes need composite reasoning (as \texttt{Comp}) or not (as \texttt{Sum}). Further, it also implies that our \model is more suitable for visual grounding task as our assembler is aware of visual cues by the Gumbel-Softmax training strategy.

\subsection{Comparison with State-of-the-Arts}
\noindent\textbf{Settings}.
We compared \model with other state-of-the-art visual grounding models published in recent years.
According to whether the model requires language composition, we group those methods into:
1)~Generation based methods which select the region with the maximum generation probability: \textbf{MMI}~\cite{mao2016generation}, \textbf{Attribute}~\cite{liu2017referring}, and \textbf{Listener}~\cite{yu2017joint}. 
2)~Holistic language based methods: \textbf{NegBag}~\cite{nagaraja2016modeling}.
3)~Language composition based methods: \textbf{CMN}~\cite{hu2017modeling}, \textbf{VC}~\cite{zhang2018grounding}, \textbf{AccumAttn}~\cite{deng2018visual}, and \textbf{MAttN}~\cite{yu2018mattnet}.
4)~Composition methods with external parser: \textbf{GroundNet}~\cite{cirik2018using}, \textbf{parser+CMN}, and \textbf{parser+MAttN}.
\model belongs to the fourth category, but its language composition is more fine-grained than others.
We compared with them on three different settings: ground-truth regions, detected regions, and segmentation masks.

\noindent\textbf{Results}. From Table~\ref{table:2} and Table~\ref{table:3}, we can find that:
1) the triplet composition models mostly outperform holistic models. This is because taking the advantage of linguistics information by decomposing sentences, even coarse-grained, is helpful in visual grounding.
2) Our model outperforms most triplet models with the help of fine-grained composite reasoning.
3) The parser-based methods are fragile to parser errors, leading to performance decline. However, our model is more robust because of the dynamic assembly and end-to-end train strategy.
Although some of the performance gains are marginal, one should notice that it seems \model balances the well-known trade-off between performance and explainability~\cite{hu2018explainable}. As we will discuss in the following, we achieve the explainability without hurting the accuracy.

\subsection{Qualitative Analysis}
\label{sec:4.5}
In this section, we would like to investigate the internal reasoning steps of our model by qualitative results\footnote{Since our work focuses on complex language cases, we mainly conducted qualitative experiments on RefCOCOg. More qualitative results are given in supplementary material.}. In Figure~\ref{fig:5}, we visualize the tree structures, the module assembly, the attention map at each intermediate step, and the final results. In Figure~\ref{fig:6}, we visualize the reasoning process inside \texttt{Comp} modules.
With these qualitative visualizations, we can have the following observations:
1) The visual concept words usually are assembled by \texttt{Sum} module while the relationship concept words are usually assembled by \texttt{Comp} module.
2) The attention maps of non-visual leaf nodes, \textit{e.g.}, `directly' in \ref{fig:5}(d), are usually scattered, while visual ones, \textit{e.g.}, `girl' in \ref{fig:5}(d), are usually concentrated.
3) \texttt{Comp} modules are aware of relationships, \textit{i.e.}, it can move the attention from the supporting objects to the target objects, as shown in Figure~\ref{fig:6}.
4) Along the tree, attention maps become more sharp, indicating the confidence of our model become stronger.

All the above observations suggest that our \model can reason along the tree and provide rich cues to support the final results.
These reasoning patterns and supporting cues imply that our model is explainable.
Therefore, to further investigate the explainability of our model, we conducted a human evaluation to measure whether the internal reasoning process is reasonable.
Since the state-of-the-art model MAttNet~\cite{yu2018mattnet} does not contain internal reasoning process but only sums up three pre-defined module scores which directly point to the desired object, we compared with AccumAttn~\cite{deng2018visual} for it performs multi-step sequential reasoning and has image/textual attention at each time step.
We first presented 60 examples with internal steps of each model to 6 human evaluators, and asked them to judge how clear that the model was doing at each step.
Then each evaluator rated each example on 4-point Likert scale~\cite{likert1932technique} (unclear, slightly clear, mostly clear, clear) corresponding to scores of 1, 2, 3, and 4.
The percentage of each choice and average scores are shown in Figure~\ref{fig:7}.
We can find that our model outperforms AccumAttn~\cite{deng2018visual} and is often rated as ``clear''. It indicates that the internal reasoning process of our model can be more clearly understood by humans.

\section{Conclusion}
In this paper, we proposed Neural Module Tree Networks (\modelend), a novel end-to-end model that localizes the target region by accumulating the grounding confidence score along the dependency parsing tree of a natural language sentence. \model consists of three simple neural modules, whose assembly is trained without additional annotations. Compared with previous visual grounding methods, our model performs a more fine-grained and explainable language composite reasoning with superior performance, demonstrated by extensive experiments on three benchmarks.

{\small
\noindent\textbf{Acknowledgements.}
This work was supported by the National Key R\&D Program of China under Grant 2017YFB1300201, the National Natural Science Foundation of China (NSFC) under Grants 61622211 and 61620106009, the Fundamental Research Funds for the Central Universities under Grant WK2100100030, and partially by NTU Data Science and Artificial Intelligence Research Center (DSAIR) and Alibaba-NTU JRI.

\bibliographystyle{ieee_fullname}
\bibliography{citations}
}

\onecolumn
\appendix
\begin{center}
\bf \Large Supplementary Material for ``Learning to Assemble Neural Module Tree Networks for Visual Grounding''
\end{center}

\section{Implementation of Tree LSTM}
We simplified the implementation of tree LSTM (Eq.~\ref{eq:4}) in the main paper as:
\begin{equation}
    \bm{c}_t^{\uparrow}, \bm{h}_t^{\uparrow} = \mathop{\textrm{TreeLSTM}}(\bm{e}_t, \{\bm{c}_{tj}^{\uparrow}\}, \{\bm{h}_{tj}^{\uparrow}\}),\quad j\in\mathcal{C}_t,
\end{equation}
where $\bm{c}_{tj}^{\uparrow},\, \bm{h}_{tj}^{\uparrow}$ denote the cell and hidden vectors of the $j$-th child of node $t$.
Specifically, our tree LSTM transition equations are:
\begin{equation}
    \tilde{\bm{h}}_t = \sum\nolimits_{j \in \mathcal{C}_t} \bm{h}_{tj}^{\uparrow},
\end{equation}
\begin{equation}
    \bm{i}_t = \sigma(W^{(i)} \bm{e}_t + U^{(i)} \tilde{\bm{h}}_t + \bm{b}^{(i)}),
\end{equation}
\begin{equation}
    \bm{o}_t = \sigma(W^{(o)} \bm{e}_t + U^{(o)} \tilde{\bm{h}}_t + \bm{b}^{(o)}),
\end{equation}
\begin{equation}
    \bm{u}_t = \tanh(W^{(u)} \bm{e}_t + U^{(u)} \tilde{\bm{h}}_t + \bm{b}^{(u)}),
\end{equation}
\begin{equation}
    \bm{f}_{tj} = \sigma(W^{(f)} \bm{e}_t + U^{(f)} \bm{h}_{tj}^{\uparrow} + \bm{b}^{(f)}),
\end{equation}
\begin{equation}
    \bm{c}_t^{\uparrow} = \bm{i}_t \odot \bm{u}_t + \sum\nolimits_{j \in \mathcal{C}_{t}} \bm{f}_{tj} \odot \bm{c}_{tj}^{\uparrow},
\end{equation}
\begin{equation}
    \bm{h}_t^{\uparrow} = \bm{o}_t \odot \tanh(\bm{c}_t^{\uparrow}),
\end{equation}
where $\odot$ is the element-wise multiplication, $\sigma(\cdot)$ is the sigmoid function, $W$, $U$, $b$ are trainable parameters.

\section{More Qualitative Results}
\begin{figure*}[htb]
    \centering
    \includegraphics[width=0.95\linewidth]{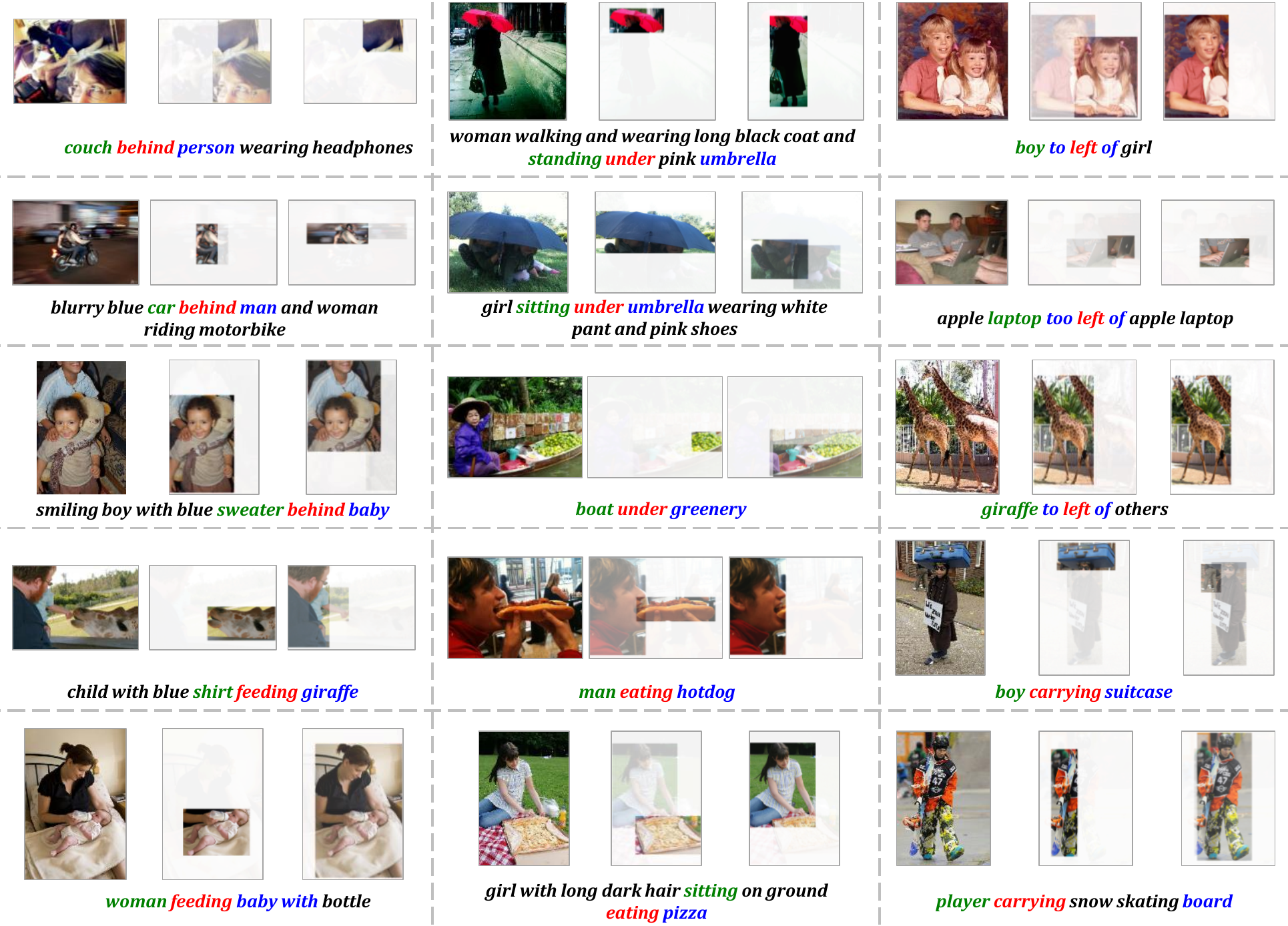}
    % \vspace{-5pt}
    \caption{The compositional reasoning inside \texttt{Comp}. Each example contains the original image (left), the contextual attention map (middle) and the output attention map (right). We represent partial tree structure by colors: red for the current node, blue for children and green for parent.}
    % \vspace{-10pt}
    \label{fig:8}
\end{figure*}

\begin{figure*}[t]
\centering
\subfigure[correct]{
\includegraphics[width=0.95\linewidth]{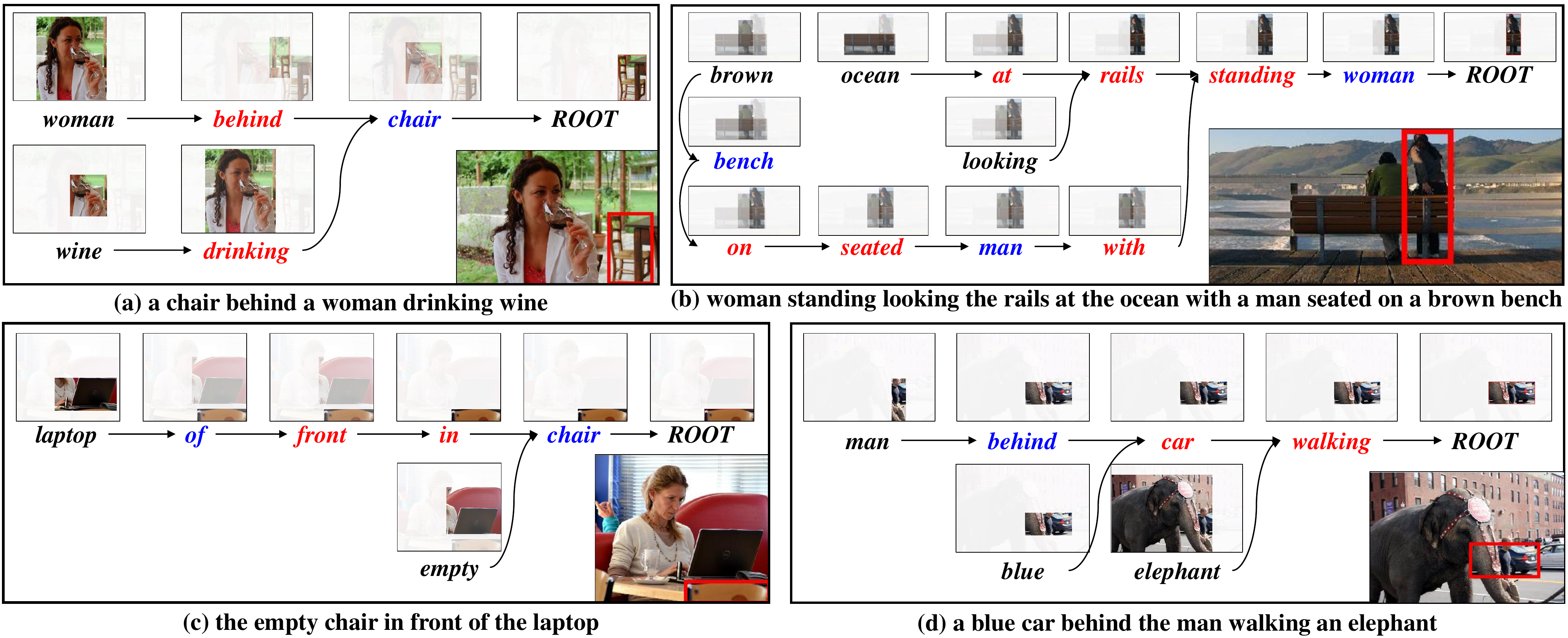}
\label{fig:9a}
}%

\subfigure[incorrect]{
\includegraphics[width=0.95\linewidth]{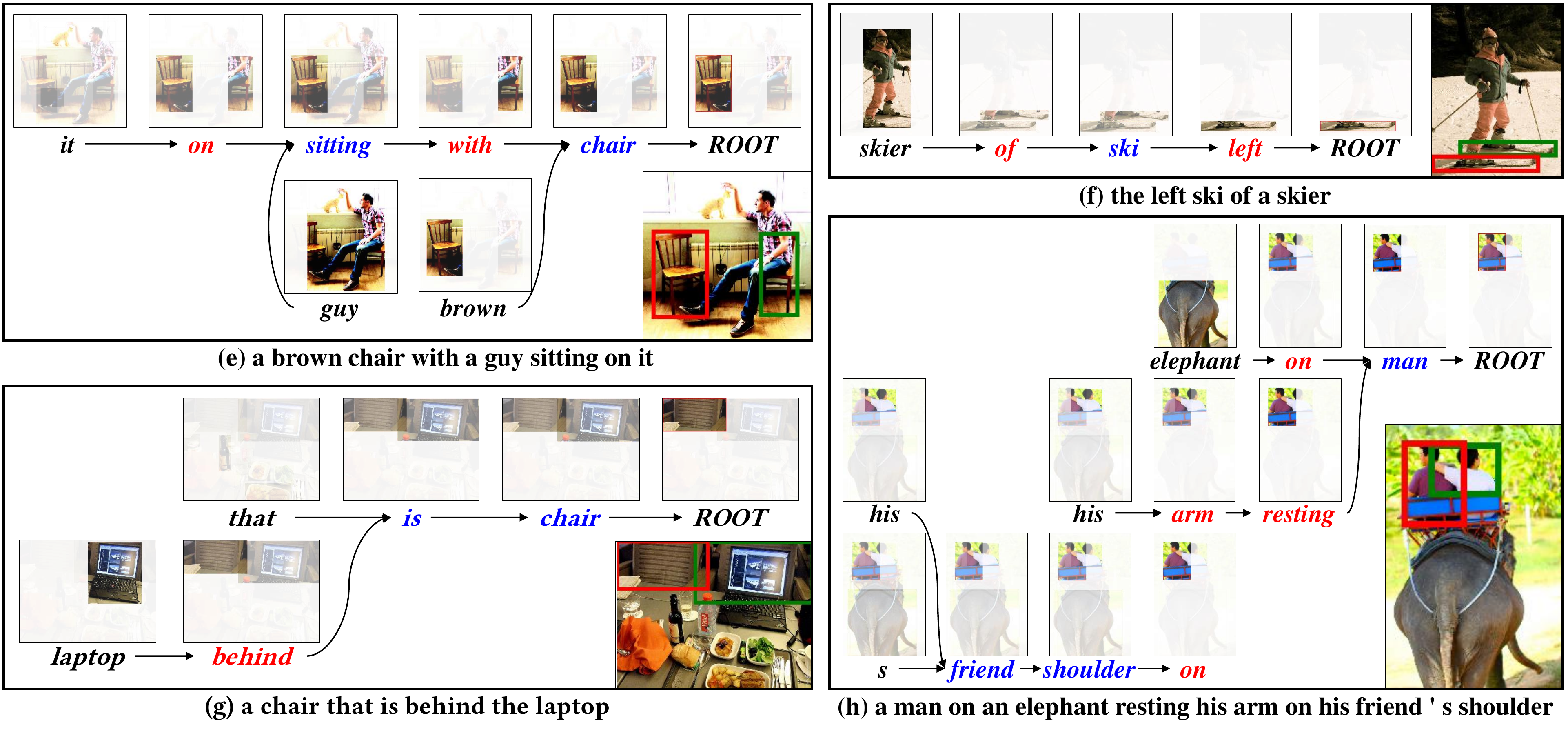}
\label{fig:9b}
}%
% \vspace{-3pt}

\centering
% \vspace{-10pt}
\caption{Qualitative results with ground-truth bounding boxes. Words in different colors indicate corresponding modules: black for \texttt{Single}, red for \texttt{Comp}, and blue for \texttt{Sum}. The bottom right corner is the original image with a green bounding box as ground-truth and a red bounding box as our result.}
% \vspace{-10pt}
\label{fig:9}
\end{figure*}

\begin{figure*}[htb]
    \centering
    \includegraphics[width=0.95\linewidth]{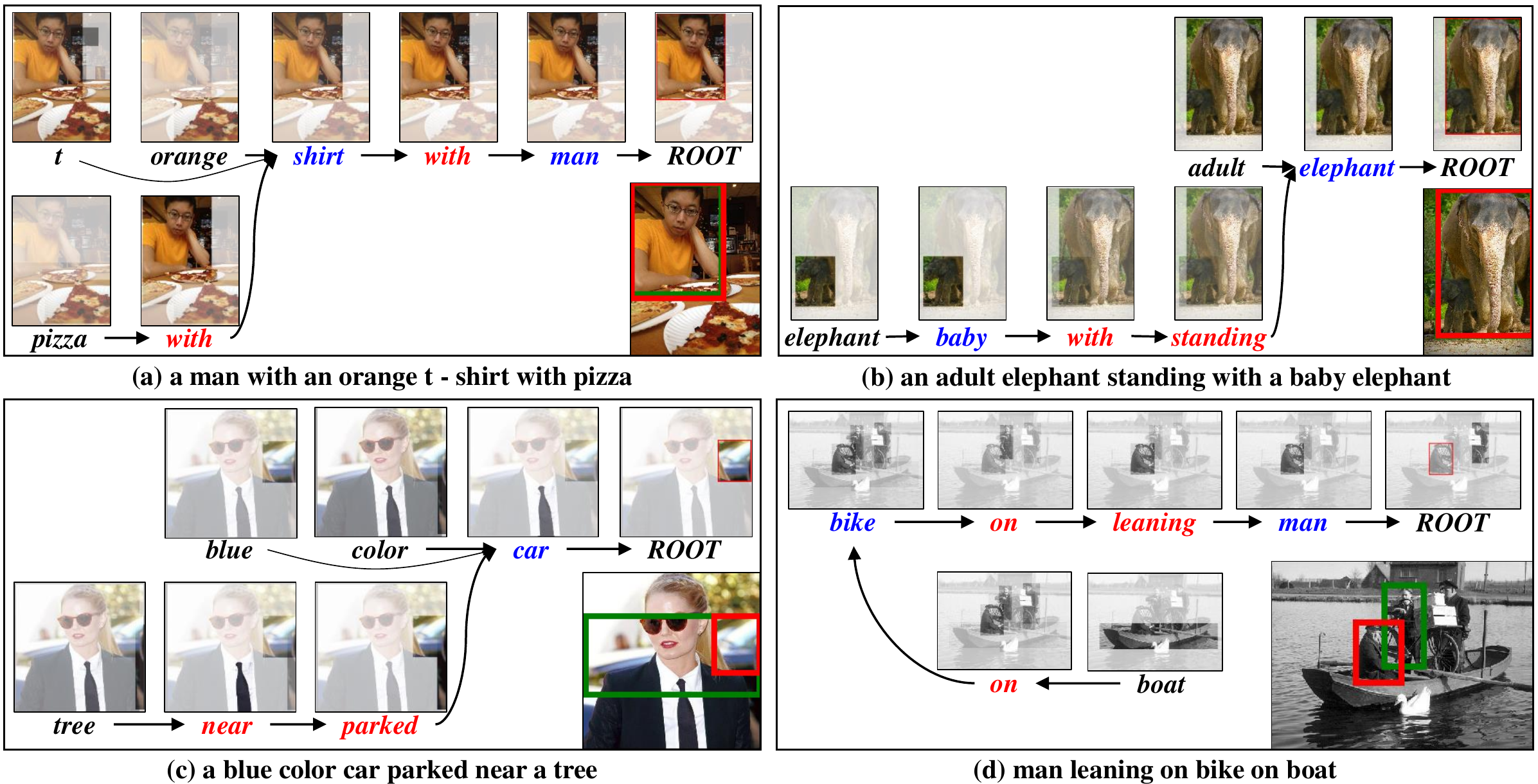}
    % \vspace{-5pt}
    \caption{Qualitative results with detected bounding boxes. Words in different colors indicate corresponding modules: black for \texttt{Single}, red for \texttt{Comp}, and blue for \texttt{Sum}. The bottom right corner is the original image with a green bounding box as ground-truth and a red bounding box as our result.}
    % \vspace{-10pt}
    \label{fig:10}
\end{figure*}

\begin{figure*}[htb]
    \centering
    \includegraphics[width=0.95\linewidth]{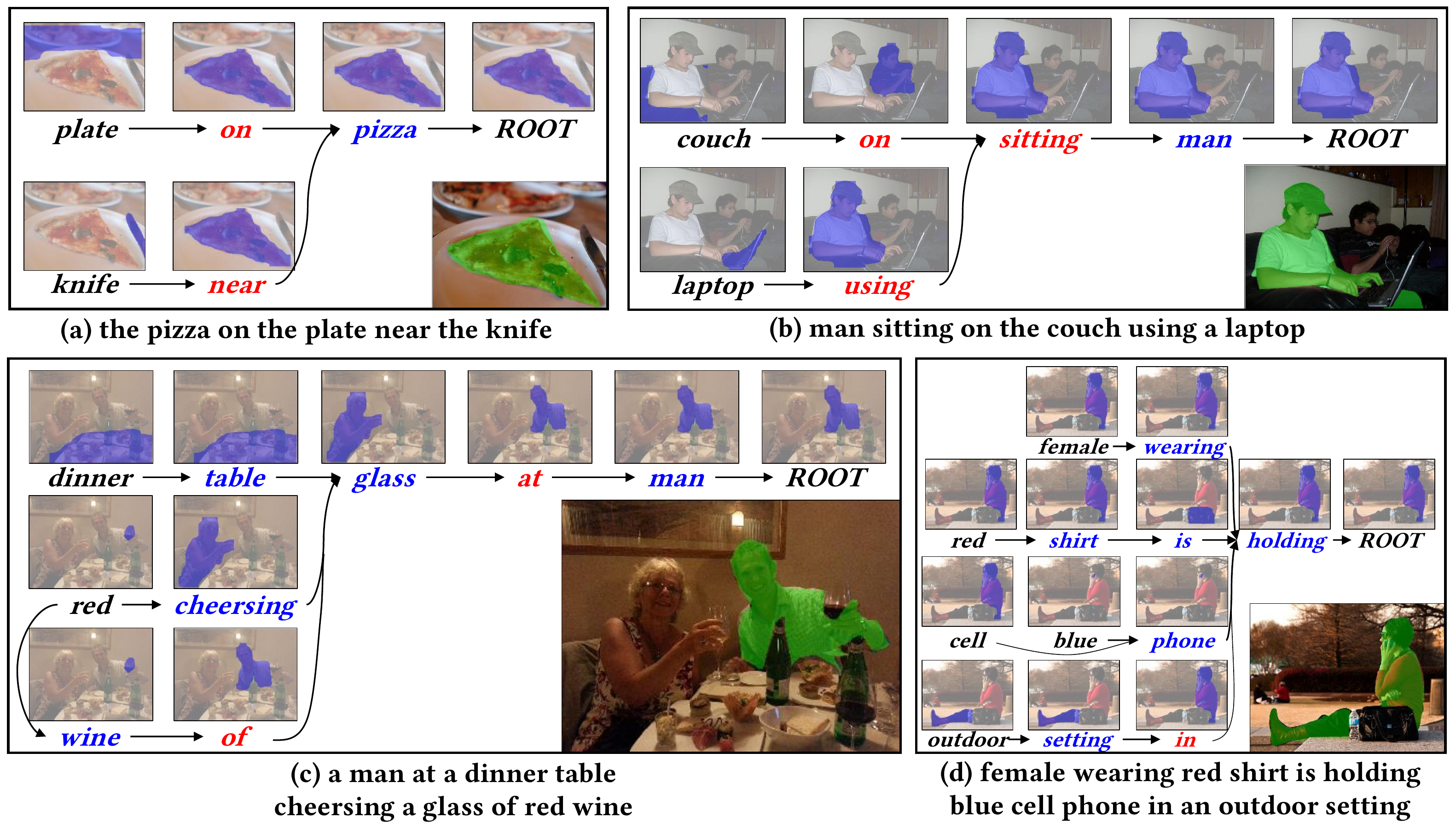}
    % \vspace{-5pt}
    \caption{Qualitative results with detected masks. Words in different colors indicate corresponding modules: black for \texttt{Single}, red for \texttt{Comp}, and blue for \texttt{Sum}. The blue masks indicate the regions with maximum score, and the green masks indicate the ground-truth.}
    % \vspace{-10pt}
    \label{fig:11}
\end{figure*}

In this section, we provide more qualitative results to demonstrate the internal reasoning steps of \textsc{NMTree}.
In Figure~\ref{fig:8}, we visualize the reasoning process inside \texttt{Comp} modules.
In Figure~\ref{fig:9}, Figure~\ref{fig:10}, and Figure~\ref{fig:11}, we visualize the tree structures, the module assembly, the attention map at each intermediate step, and the final results.
Specifically, Figure~\ref{fig:9} are qualitative results with ground-truth bounding boxes. As comparison, we also show some failure cases.
Figure~\ref{fig:10} are qualitative results with detected bounding boxes.
Figure~\ref{fig:11} are qualitative results with detected masks.

\end{document}